# Reinforcement Learning Based Safe Decision Making for Highway Autonomous Driving

Arash Mohammadhasani[1], Hamed Mehrivash[1], Alan Lynch[1], Zhan Shu[1]

*Abstract*— In this paper we develop a safe decision-making method for self-driving cars in a multi-lane, single-agent setting. The proposed approach utilizes deep reinforcement learning (RL) to achieve a high-level policy for safe tactical decision making. We address two major challenges that arise solely in autonomous navigation. First, the proposed algorithm ensures that collisions never happen, and therefore accelerate the learning process. Second, the proposed algorithm takes into account the unobservable states in the environment. These states appear mainly due to the unpredictable behavior of other agents, such as cars, and pedestrians, and makes the Markov Decision Process (MDP) problematic when dealing with autonomous navigation. Simulations from a well-known self-driving car simulator demonstrate the applicability of the proposed method.

## I. INTRODUCTION

### A. Motivation

The problem of autonomous navigation has been the focus of research over decades[1]. In recent years there have been even more interests in self-driving cars. Developing these autonomous creatures has been an active area of research [2,3,4] due to its high potential to generate more efficient road networks that are much safer for the passengers.

Endowing a car with the ability to make tactical decisions, referred to as "Driving Policy", is key for enabling fully autonomous driving. The process of observing, i.e., forming a model of the environment, consisting of location of all dynamic and stationary objects (such as cars, pedestrians, and so forth) is well defined. While sensing is well understood, the definition of Driving Policy, its critical assumptions, and its functional safety are less studied [5].

Although there has been considerable progress in the last decade [6], driving policies for fully autonomous navigation is still an open problem, specifically in uncertain, dynamic environments surrounded by other agents. The challenges arise mainly because the other decision makers' goals and desired routes are not clear to the planning system, and, communication of such policies is not practical because of physical restrictions. These problems stimulate the exploitation of decentralized collision avoidance methods [7].

In this paper, attention is focused on the tactical decision making, and top-level decision are proposed to adjust the actions of the vehicle to the latest traffic situation such that some criteria of driving (safety, quickness, consistency) are satisfied. Specifically, these decisions control whether to change lanes, which lane to go and whether accelerate or not.

In the authors mind, there are three types of methods to deal with high-level decision making for autonomous navigation. The first group include rule-based methods, which are successfully implemented in the DARPA Urban Challenge [2,8]. However, a drawback of such methods is that they lack the capability to be used in unknown situations, making it impossible to apply them in the real-world scenarios [9,10].

The second group treats the decision-making problem as a motion planning task, see [11] and the references therein where such a methodology is applied to highway and intersection scenarios. Although it has been successful in multiple cases, the sequential construction, involving the estimate of the trajectory of the cohabited vehicles and thereafter the determination of the Ego vehicle (the car under consideration) desired path accordingly, results in a defensive action which does consider the possible interactions during motion planning [9].

The third category -which is also the focus of this paper- is to use the theorems developed in RL-domain in order to propose safe decision-making algorithms that are applicable for autonomous navigation.

Reinforcement learning provides a way to learn arbitrary policies considering specific goals. In recent years learning- based approaches have been used to tackle similar or related problems, like learning-based on human driving [14], inverse reinforcement learning [15], end-to-end algorithms that map sensed inputs (mainly visual, images) directly to control signals [16, 17], and methods that understand the scene via learning to make tactical decisions [18, 19].

While offline solutions are able to handle complex scenarios and compute a policy before implementation (see [20] for an example on an intersection scenario), but they are impractical because there exists a large number of presumable real-world situations. So, it is intractable to precompute a general policy that would be applicable in all possible cases. Online methods calculate a policy while they are experiencing the world, which makes them dominant compared to than their offline counterparts.

On the other hand, the combination of reinforcement learning with deep learning (DL) is a well-known approach to obtain a human-level control (see [21]). In [22] such control scheme is illustrated on games using the combination of Q-Learning and neural networks, in which RL handles the planning part, while DL is used as a solution to the representation learning.

Motivated by these recent statements, the focus of this paper is to propose an appropriate online decision-making algorithm for fast yet safe autonomous navigation by DQN networks. We will implement our approach in the simulator designed by [23]. Indeed, we neither investigate the vehicle/environment model, nor it's similarity to real-world scenarios. Instead we concentrate on developing an appropriate policy given the specific model/task which will be tested on this environment.

### B. Related Works

We split this part into three sub-sections, the first one deals with finding an optimal solution that addresses the RL-based autonomous navigation along with a brief review on the sample efficiency of the Q-learning algorithm. The second part investigates environment and state modeling part of autonomous navigation, and the last part is an overview of available safe self-driving approaches in the literature of RL-based autonomous navigation. In later sections, we will justify this point of view to the problem in hand.

[1]Department of Electrical and Computer Engineering, University of Alberta, Edmonton, Canada, T6G 2V4{Arash.mhasani , h.mehrivash , Alan.lynch , zshu1}@ualberta.ca.

*1) Q-learning for Autonomous Navigation*

In spite of different views to the self-driving task in terms of assigning the reward function, state modeling, and selection of the environmental model, it seems that Q-learning is the workhorse of policy optimization for the autonomous navigation. In [24], DQN is applied in a multi-lane, multi-agent setting with single convolution layer with rectifier activation function (ReLU) to train an agent to exit a highway in the shortest possible time. In [25] the DQN is also applied, training the EGO car to drive as fast as possible but this time in different settings; with two hidden layers, 100 neurons each, and exponential activation function. In [26], the same procedure is applied where they investigated the performance of the proposed network in different phase of the training (initial, middle and final stage of policy optimization). In [27] the DQN is applied but this time with recurrent network architecture to take into account the effect of the unobservable states.

Q-learning, although shown to be practical, may require many samples to learn [36,37]. The controversial part is that the epsilon-greedy policy optimization scheme (see [12] page 153) may not be that efficient. In [35] it is proved that Q-learning with UCB exploration achieves considerably less regret than its epsilon-greedy counterpart. In [38] the noisy-nets idea is introduced, and it is claimed that it outperforms epsilon-greedy by 48% in the sense of mean human-normalized scores.

In this paper we address this problem with a different methodology, in fact we will show that the available Q-learning with epsilon-greedy action selection scheme, can be made much more efficient by excluding unsafe state-actions.

Other works inspired by policy gradients approaches can be found in [28, 29].

*2) State Modeling*

For the state modeling part, the authors noticed that there are two distinct approaches in the context of RL-based tactical decision-making for autonomous navigation.

The first group models the problem in the standard MDP framework. In this case, the states generally include velocities and positions of the other agents/vehicles in the environment relative to the Ego car (and the absolute velocity of the Ego itself), this kind of state representation can be found in [10, 24, 26].

However, there is a second approach which arises from the fact that the problem is indeed a multi-agent scenario. This kind of approaches try to consider the states that partly represent the decision of the other agents in the environment, an extensive analysis of such approaches and their advantages can be found in the recent work [7]. The common tool to handle the unobservable states is to maintain a belief for them and further propagate that belief in time, see [9, 27, 30, 31] as examples, where Monte Carlo tree search and recurrent networks are applied to address the non-observable problem in hand.

*3) Reward Shaping*

The third grouping coincides with assigning different reward function to the problem. While minimum travel time seems to be a common choice in designing the reward function, the collision avoidance reward is the controversial part.

The first category is similar to what has been done in [26, 32], where a high negative reward is assigned in order to avoid possible collisions. Although this is a common practice in RL tasks, collision must never happen in real-life autonomous navigation scenarios. Motivated by this statement, [9, 10, 24] utilized another approach (Q-masking) to ensure that collisions never happen. This treatment is to exclude those actions that will result in possible collision at the final step of policy optimization.

The third category tries to model the safety as a restriction over actions and states and connects them to the domain of constrained MDPs (see [41-43]). A comprehensive survey on these methods can be found in [40].

We will further analyze the methods (3) mentioned above in section "SAFE DECISION-MAKING".

*B. Contributions*

In this paper a DQN-based approach is developed for safe tactical decision-making on highway environments. More specifically, we present a novel approach that is able to

- Achieve minimum travel time (highest average velocity) while avoiding collisions at the same time. We investigate the sample efficiency of the proposed method by excluding those unsafe actions, and stop our DQN network from learning those unsafe state-action pairs.

- Make safe decisions which is also robust against possible unpredicted behavior of other agents. Thorough this, we consider the worst-case scenario when the agent makes the decisions, we found this method much more practical than classifying the problem as conventional Partially Observable Markov Decision Process assumption (POMDP).

- We will also show that Q-masking algorithm proposed in the literature is suboptimal solution to the safe decision-making.

Although we classified above statements as distinct contributions (as indeed they are different problems), we will introduce a unified approach to handle them We refer to this method as " robust Q-masking".

To summarize the above contributions, we provide a dedicated Q-learning version for the problem of autonomous navigation.

The remainder of this paper is organized as follows: Sec. II introduces the RL problem statement. In Sec. III, we formulate the RL-based autonomous navigation problem.

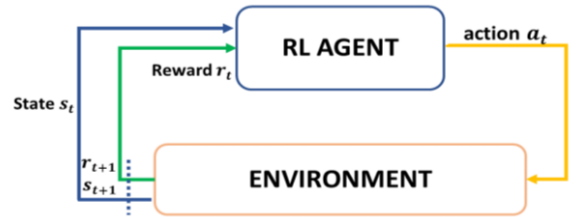

Figure 1. Based on the the state s_t at current time t, the agent performs the action a_t which leads to a new state s_(t+1). The agent also receives a reward r_(t+1) for the selected action.

Section IV describes the kinematic model of the environment while Section V investigate the existing methods of safe decision making in RL domain, along with a discussion on each. Section VI discusses the proposed safe & robust decision-making approach. Finally, Section VII describes the simulator settings and the corresponding results.

II. GENERAL RL PROBLEM DEFINITION

The model is developed based on Markov Decision Process (MDP) framework, which is a set of $(S, A, \gamma, P(s', r|s, a), R)$ where: a) Set of Environment states $(S)$, b) Set of Actions $(A)$, c) Discount Factor $(\gamma)$,  d) Reward $(R)$,  e) State Transition Probabilities $(P(s', r|s, a))$ [12].

The RL agent and the environment interaction is in discrete time sequence t = 0,1, 2,… (see Fig. 1, [2]). At each time instant "t", and based on received information about the environment's state $s_t \in S$, the RL agent decides an action $a_t \in A(s_t)$. One-time step later, as a

result of the action performed in $s_t$, the RL agent receives a reward

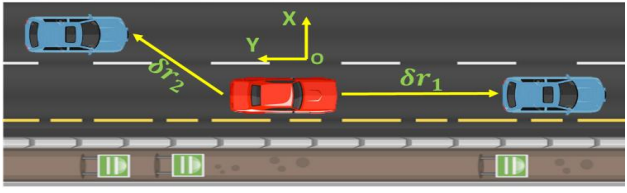

Figure 2. The environment of the agent includes two lanes with multiple leading and following vehicles, which are controlled by an expert system.

$r_{t+1} \epsilon R$, and set foot on a new state $s_{t+1}$.

A policy $\pi$ is a map $S \rightarrow A$ that at each state $s$ selects an action to be performed when in state $s_t$. The goal of the RL agent is to find a policy to maximize total expected reward.

We define state-action value function as the expected return starting from s, taking the action a, and thereafter following policy $\pi$, [12]. The value function is the expected sum of the discounted rewards as follows:

$$Q_\pi(s,a) = E_\pi\{R_t|s_t = s, a_t = a\}$$
$$= E_\pi\left\{\sum_{k=0}^{k=\infty} \gamma^k r_{t+k+1}|s_t = s, a_t = a\right\} \quad (1)$$

The discount factor $\gamma$ is in [0,1], [12, p. 55]. The state-action value function associated with the optimal policy $\pi^*$ is called optimal state-action value function $Q^*$, and defined as:
$$Q^*(s,a) = max_\pi Q_\pi(s,a), \forall\, s\, \epsilon\, S, \forall\, a\, \epsilon\, A \quad (2)$$

*Table 1*

| BATCH DEEP-Q-LEARNING [13] |
|---|
| Collect a set of interaction samples $D: \{(s_i, a_i, s_{i+1}, r_{i+1})\}$ |
| Initialize state-action value function $Q(s, a)$ with random parameters $\theta$ |
| Loop |
| Sample random mini-batch of transitions $s_i, a_i, s_{i+1}, r_{i+1}$ from $D$ |
| Set $target_j = \begin{cases} r_j & s_{i+1}: \text{ terminal} \\ r_j + \gamma max_{a'} Q(s_{i+1}, a'; \theta) & s_{i+1}: \text{ non}-\text{terminal} \end{cases}$ |
| Perform a gradient descent step on $\left(target_j - Q(s_i, a_i; \theta)\right)^2$ with respect to $\theta$ |
| Until termination conditions are satisfied. |

### III. PROBLEM STATEMENT

We consider the problem of autonomous lane changing in a multi-lane single-agent setting and set up the problem in the MDP environment that we mentioned in the last section. In particular, we define the set of states, actions, then we discuss pertinent reward functions. Given these settings we propose an appropriate algorithm to handle tactical decision-making in autonomous navigation.

#### A. States

We assume that the agent in Fig.2 can observe four nearby vehicles. By "nearby vehicles", we refer to those with minimum distances to the Ego car. Let us introduce the reference coordinate system by $O_{XY}$ (which is visible by axes in yellow and letters in green). The states then include relative positions (denoted by $\delta r_i$) of the nearby cars to the Ego car, relative velocities, and heading angles of the vehicles and absolute velocity and heading of the Ego car, all expressed in reference coordinate systems.

We also consider a flag indicating that the corresponding vehicle is currently present or not, in order to detect the times that less than four cars are visible.

To summarize, the states are:

- Relative distances ($\delta r_i$),
- Relative velocities ($\delta v_i$)
- Presence flag (flag)
- Heading angle ($\psi_i$)

For each of four surrounding vehicles. And
- Velocity ($V_{ego}$)
- Heading angle ($\psi_{ego}$)

For the Ego car.

So, at any instant "t", the state vector will be:
$$S = \{\delta r_i, \delta v_i, \psi_i, \text{flag}_i, V_{ego}, \psi_{ego}\}_{i=1,2,3,4} \quad (3)$$

#### B. Actions

As we mentioned earlier, we only consider high-level decision-making actions. Table 2 summarizes the available set of actions in every state. These are the minimal set of actions to achieve the final objective of the autonomous navigation (i.e., minimum travel time while avoiding collision).

Considering the fact that other agents/vehicles are controlled by an expert system (IDM system [23]), there is always at least one action which is safe.

*Table 2*

| Available Actions | Descriptions |
|---|---|
| $a_0$ | Stay in the current lane (IDLE) |
| $a_1$ | Change lanes to the right |
| $a_2$ | Change lanes to the left |
| $a_3$ | Faster |
| $a_4$ | Slower |

It should be noted that for actions $a_3$ & $a_4$, they are only allowed while the minimum/maximum velocity is not violated.

So, at any instant "t", the $A$ set will be:
$$A = \{a_0, a_1, a_2, a_3, a_4\} \quad (4)$$

#### C. Rewards

The assignment of an appropriate reward function is a critical task in RL-based autonomous navigation, and usually reflects different behaviors of the agent (e.g., driving in the rightmost lane, driving consistently; changing the velocity in a smooth way). But the two most important criteria are driving as fast as possible and collision-avoidance. Possible reward functions along with a discussion on each is presented in the next section.

### IV. SAFE DECISION-MAKING

In this section, we consider the safe decision making on a given MDP. One reason for this (beside that the safe decision making is the focus of this paper) is that this topic is usually less discussed in the conventional RL references, mainly because they assume that the MDP environment is not known a priori (see cliff-world example in [12], page 132), and then they try to achieve the best performance given this unknown environment. However, this is not the case in RL-based autonomous navigation since at any moment the safe decision boundary can be determined by means of kinematic information from nearby vehicles (we will show this shortly).

There are two main reasons that motivate us to focus on the safety aspect of the decision-making process:

Firstly, safety is the most necessary requirement in autonomous navigation. Indeed, collisions must be avoided under any circumstances, and this is not a trivial task. Applying a reward function which penalizes collisions with huge negative reward, the agent will learn appropriate behaviors after some epochs of its exposure to training data (for now, forget about the sample-efficiency criterion), and then the learned model will be employed in real-world scenarios (or is tested using some unseen data) The main problem arises when the agent encounters a scenario which it has not been exposed to in the training data-and there millions of new situations for example by just adding a pedestrian to some previous data-, so it cannot evaluate the safety level of the new situation, and consequently its decision may result in collisions. Although this may not be critical problem in common RL-tasks (e.g., a robot collecting cans), but it is a matter of life and death in autonomous navigation.

Secondly, by using this safety information we can heavily simplify the reward function and therefore make the overall learning process faster and more sampling-efficient. The main justification behind this statement is that we will lose nothing by stopping our agent from learning those unsafe state-action pairs, because the optimal decision-making strategy will never select any action which leads to collision.

The same arguments also hold for the unobservable states. Consider an agent which always makes decisions with given states (introduced in Section III). What is the guarantee that other agents maintain the same state during the period of two consecutive epochs? In other words, if we make safe decision based on the current velocity of the frontier car, what is the guarantee that the car (which can be seen as another agent) will not suddenly brake with maximum possible negative acceleration? What is the guarantee that the agent in the adjacent line will not suddenly drive into the agent's line?

To better depict these unobservable states, visit Fig.5.

Now that we have justified the necessity of safe-decision making algorithm and inclusion of unobservable states, we will focus on the methods to address them. The reason that we are introducing these methods here in details (and not just briefly in Introduction section), is that we will use them later as baseline methods to compare our results with, so it is beneficial to know them first.

Also note that these methods are not necessarily different in Q-learning methodology, however we refer to them with different "Q-…" names, as this is common practice in the literature. They are also sorted in a sequence that they can act efficiently (in author's point of view).

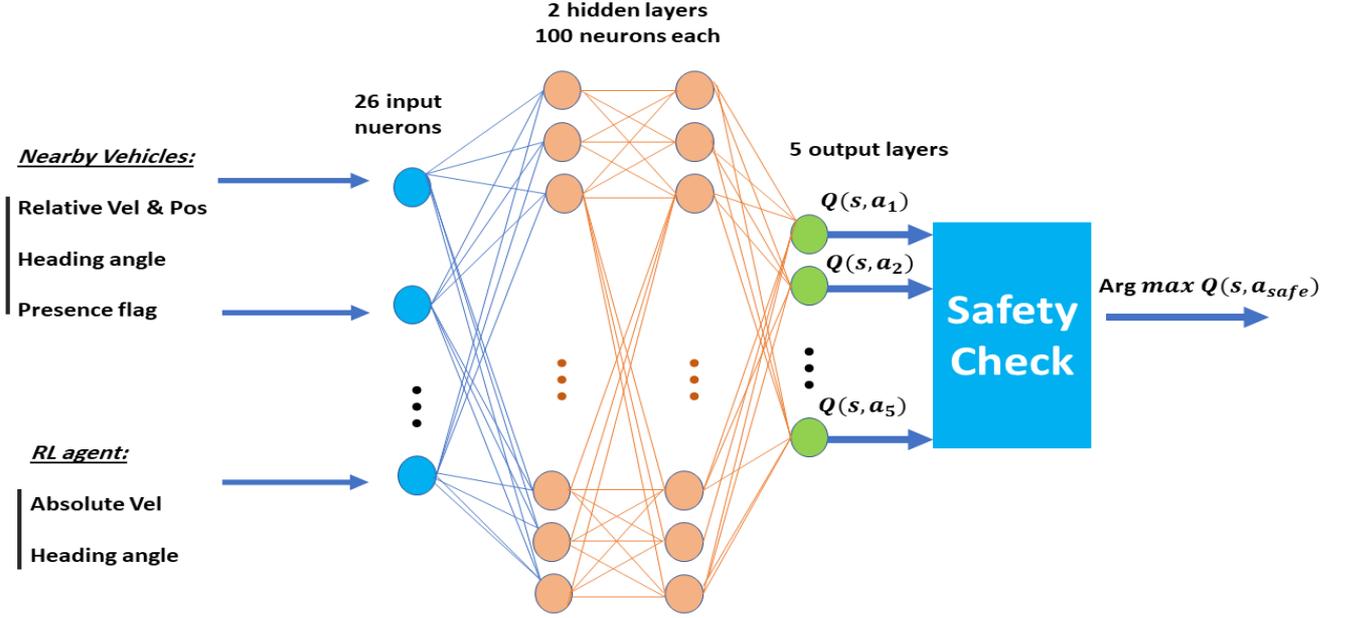

Figure 3. The architecture of the DQN network of Q-masking algorithm, inputs are the new state and outputs are the estimated Q values of the five actions. Before executing each action, the safety check algorithm verifies the safety of the selected action Q [10].

*1) Traditional Q-Learning*

This is a common practice in applying RL algorithms in autonomous navigation. The reward function is chosen as:

$$R_{1,2}(t) = b \frac{v-v_{min}}{v_{max}-v_{min}} - c \times collision \quad (6)$$

Where "$b$", and "$c$" are some constant coefficients (hyperparameters) and takes into account the minimum travel time and safe decision-making accordingly.

*2) Constrained Q-Learning*

In these methods, the safety is modeled by restricting the state-action space. In such constrained MDPs, the policy optimization method will be modeled by:

$$Q_{t+1}(s_t, a_t) = Q_t(s_t, a_t) + \alpha_t(r_{t+1} + \gamma max_{a'}Q_t(s_{t+1}, a') - Q_t(s_t, a_t))$$

Subject to:
$$\begin{cases} c_i(s_t, a_t) = C_i & i = 1, 2, \dots, k' \\ c'_i(s_t, a_t) \leq C'_i & i = k'+1, \dots, k \end{cases} \quad (7)$$

Where $c'_i$ and $c_i$ are inequality and equality constraints over states and actions, which in general can be nonlinear (see Eq. 11 & 12), and $C_i$ and $C'_i$ determine the boundary of the feasibly set for the states and actions (hyper parameters).

A general treatment for these constrained policy optimization problems is to 1) linearization of the nonlinear constraints and then 2) adding them to the original cost function by Lagrange multipliers. Then the overall problem is to learn the optimal policy along with Lagrange multipliers which satisfy the constraints (to see the constraints, see Eq. 11 & 12), for details see [40-43].

However, the problems with applying these methods in safe autonomous navigation is that 1) many episodes are required to learn these newly added Lagrange multipliers (constrained area) particularly because these constraints are highly nonlinear in the case of safe autonomous navigation 2) these learned constraints cannot easily be generalized (e.g., when the number of lanes increases), so the constrained optimization problem must be solved again for the new scenarios.

We will demonstrate these weaknesses further in simulation part.

*3) Q-Masking*

The idea behind this method is simple, but really fruitful in the case of RL-based autonomous navigation. The argument is that, if we could somehow know about the safety of our actions a priori, why shouldn't we use that information to stop our agent from having collisions?

from onboard sensors (e.g., cameras, lidars, etc.) on the Ego vehicle, or that they can be transmitted to the Ego vehicle by nearby cars or some central traffic agency (for now, let's ignore the transmission delays and other related issues that may arise during transmission or inspection).

Here, is a very brief review of the method suggested in [10].

According to traffic safety rules [34], the driver must keep a safe distance from other nearby agents in order to avoid collision during abrupt braking's of the leading vehicle. This safe distance involves the dynamics of the agent and the other adjacent vehicles.

The future position of a vehicle can be described as follow:

$$P(t) = P(0) + vt + \frac{1}{2}at^2 \qquad (8)$$

Where, $P(0)$ is the current position of the vehicle at time $t = 0$. The safe distance that the Ego vehicle needs to keep from the leading vehicle is:

$$P_{ago} < P_l - P_{safe} \qquad (9)$$

Where, $P_l$ is the position for the leading vehicle.

While, the distance from the following vehicle is as:

$$P_f + P_{safe} < P_{ego} \qquad (10)$$

As a result, we have the ***free space*** $S^t$ of the agent in a lane defined by combining (3) and (4). The free space is easily calculated by removing the unsafe space from whole feasible area (see Fig. 4):

$$S^t = \{P \epsilon \mathbb{R} | P_f + P_{safe} < P < P_l - P_{safe}\} \qquad (11)$$

The controversial part however, is how to calculate these $P_{safe}$ to account for sudden braking/accelerating of other nearby vehicles. This is done in [10] & [33] by taking the maximum possible acceleration of the vehicles into account.

Finally, the safe distance can be calculated as follows [10]:

$$P_{safe} = \frac{(v_l - v_{ego})^2}{2(|a_{max,l}| - |a_{max,ago}|)} \qquad (12)$$

Here, $a_{max}$ is the maximum absolute acceleration of each Vehicle.

It is worth mentioning that for notational simplicity, here we referred to the vehicle as a single point rather than three-dimensional object, and however, the method can be easily developed to that case [33].

The idea of Q-masking is depicted in Fig.3, as it is noticeable from the figure, the main drawback associated with Q-masking is that it leaves the unchosen unsafe state- action pair unchanged. For example, consider that at any moment one of the unsafe state-action pairs seems to be the best (has the highest "Q"), as it is noticeable from Fig.3, the Q-masking method will not choose such action, but it leaves this *apparently* best state-action pair unchanged, so in the next epoch this state-action pair will generalize to adjacent states and make the agent to move toward this unsafe state-action, this sequence although will not result in collision at the end but will cause the method to be suboptimal in the sense of sample efficiency (see the simulation section).

Kinematic model of the environment to calculate the safe area ($P_{safe}$), is presented in the Appendix.

Fig. 4 represents the safe area around the Ego car in the Q-masking method.

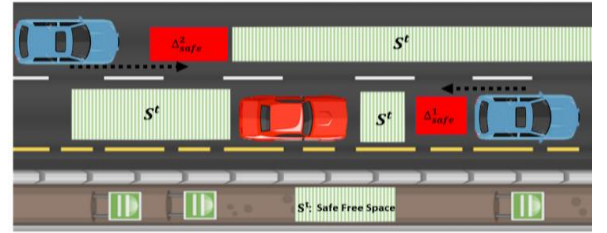

Figure 4: Safe Space calculated by Eq. 11 from current states

*4) Robust Q-Masking*

This section involves the main contribution proposed in this paper, which includes modifications to current Q-masking algorithm

*a) Generalization of Unsafe State-Action Pairs*

In the first step we need to generalize between unsafe state-action pairs. This might be trivial at first, but as we will see in simulations, it will highly speed up the learning process.

Consider that we have decided to proceed with Q-masking method by evaluation of the formula (11) at each time the agent encounter a new state, given the infinite number of situations that the agent might get involved in, it will seem to be necessary to make a connection between different situations. For example, consider a situation in which the agent drives in a *2-lanes* scenario and there is another vehicle in immediate front of the agent. Obviously driving with maximum speed in a forward direction is not a safe decision, now consider another scenario in which the Ego car is in a *4-lines* scenario with some nearby vehicles at adjacent lines and a car in short forward distant of our agent, so again driving quickly to forward direction is not a reasonable choice. We call this as a generalization of unsafe state-action pairs, which speeds up the computational process of evaluation of Eq. (11) to find the safe space particularly in large problems (e.g., more lanes, vehicles, states).

*b) Heading Angle Inclusion*

Here we mention two reasons to show that why the inclusion of heading angle into safety evaluation process is not only advantageous but also necessary.

As the first reason, consider the case when there is another vehicle in imminent front of the agent. In this case the agent equipped with safety criterion in Q-masking (see (12)), only considers the possible braking of the frontier car as a worst-case scenario, so possibly it may decide to change the lane, but what is the guarantee that the other agent will not perform lane changing at the same time? See Fig.5, where a possible lane change of nearby vehicles is shown.

As for the second reason, the inclusion of this heading angle will complete our knowledge from other agents ([7]) in the case of autonomous navigation. We believe that by taking the worst case of this angle into account, we do not need any further information about the intention of other drivers/agents, and therefore we will not encounter partially observable MDP anymore.

In fact, inclusion of this angle relieves us from employing state belief for unobservable states (for state belief and POMDPs, see [9, 27, 30, 31]).

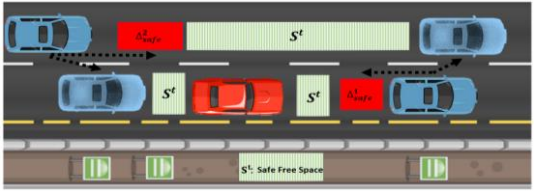

Figure 5: The safe & robust area around the RL agent along with possible maneuvers of adjacent vehicles: following/leading vehicle change lanes, vehicles merge into the safe space or the following/leading vehicle fully accelerates.

*c)   Q-Shaping*

In this section, we introduce a new paradigm to factorize the state-action value function ($Q(s,a)$) used in DQN algorithm.

Indeed, we assume the state-action value function to take the following form:

$$Q(s,a) = Q(s,a;\theta) + Q^u(s,a) \quad (13)$$

Where the first part $Q(s,a;\theta)$ is the part that will be trained through optimization process (neural networks), and the second part $\boldsymbol{Q^u(s,a)}$ takes into account the $Q$ value in unsafe state and actions.

We formulate this $\boldsymbol{Q^u(s,a)}$ as follows:
$Q^u(s,a) =$
$\begin{cases} 0, & (s,a) \notin (S^u, A^u) \\ \min(Q(s,a;\theta)) - 1, & (s,a) \in (S^u, A^u) \end{cases}$ (14)

While $(S^u, A^u)$ is the set of unsafe state-action pairs.

It should be noted that the proposed $\boldsymbol{Q^u(s,a)}$ will be updated whenever the $\boldsymbol{Q(s,a;\theta)}$ is updated.

The justification behind this idea for the shape of $\boldsymbol{Q(s,a)}$ is straightforward, whenever we are facing an unsafe state-action pair, the $\boldsymbol{Q((s^u,a^u))}$ will be less than the minimum possible amount of safe state-action pairs in $Q(s,a)$, and whenever we are updating the $\boldsymbol{Q(s,a)}$, we will also update $\boldsymbol{Q^u(s,a)}$. So this cause the unsafe-state action pairs to never be selected, and generalizes their bad (low) $\boldsymbol{Q}$ value to adjacent states (as oppose to Q-masking).

The main motivation behind this formulation is that "*if we know that some state-action pairs are unsafe a priori, why do we need to waste some time/epochs to learn them (as oppose to "Q-masking" & "constrained Q-learning" which let the neural network to be trained about these unsafe situations)?*"

Also, as we will see in the simulation part, in Q-masking method the unselected (masked) state-action pairs will maintain their assigned high reward values and this will encourage the agent to move toward them from nearby states (although not choosing them finally). The argument also holds in the case of constrained Q-learning, the question is why do we need to learn the information (unsafety of some state-action pairs) that we already have?

It is also worth mentioning that in the author's point of view, planning is not applicable in RL-based autonomous navigation framework (which is used in this paper), since the developed model of the environment will not be valid without external information (velocities & positions of the surrounding vehicles), and thus the collision-avoidance is not guaranteed anymore.

## V. Learning Algorithm

The parameters of the deep Q-learning algorithm are presented in Table3.

In this paper, we utilized the DQN with quadratic cost function:
$$e^k = \left\|\hat{Q}(k+1) - \hat{Q}(k)\right\|_2^2 \quad (15)$$
Where $\|\ \|_2$ is the 2-norm.

The architecture of the DQN network is presented in Fig.3.

We implemented the experience replay version of densely connected network of DQN, with 50 as the buffer size, 26 input dimensions (=number of states), two-hidden layers with 100 neurons each, ELU (exponential linear unit) activation function, and 5 outputs (=number of actions).

Table3 summarizes all parameters used for DQN training.

*Table 3. Training Parameters*

| | |
|---|---|
| Number of input neurons | 26 |
| Number of hidden layers | 2 |
| Number of neurons in hidden layers | 100 |
| Connection between layers | densely connected |
| Number of output neurons | 5 |
| Activation function | ELU |
| Mini-batch size | 50 |
| $\gamma$ | 0.99 |
| $\alpha$ | 0.01 |
| Optimization algorithm | Adam |
| Number of training iterations | 150 |

## VI. Implementation & Evaluation

In this section, we implement all methods that we introduced in Section IV on the environment in [23].

In terms of observation and action space the environment corresponds exactly to those we already specified in Section III, and for the reward function, it ranges from the ordinary reward in traditional-RL in Section IV, to the reward for the Q-masking and robust Q-masking.

The environment's setting includes a 4-lane road, 50 total vehicles (over 40 seconds of each episode's duration), and policy frequency of 1 second. The complete setting of the environment can be found in the Appendix.

In the remainder of this section, we first evaluate the performance of the methods in terms of sample efficiency (convergence time); this means how much it takes for each method to arrive to its steady states (plateau) regardless of the optimality of the final value. In the final step however, we will discuss the quality of this final solution (we will compare them in terms average total reward).

For the whole results in this section, the agent is first learned with experiencing different scenarios. Then, the trained agent is tested on the same setting, but this time with unseen data. The learning phase contains 200 episodes with 40 seconds as the duration for each. Then, for the test phase, 20 unseen episodes have been considered with the same duration (40s).

We tried to broaden our implementations to map the effect of different hyper parameters such that possibly bad tuning of them do not bias our judgement. Specifically, we tried different limit bounds for constrained-Q ($C_i$ and $C_i'$ in Eq.7), and different collision rewards in traditional-RL (see "$c$" in Eq.6), and different feasible thresholds for safety bounds in Q-masking and Robust Q-masking algorithms (see Fig.4 & 5).

The results provided below correspond to those hyper parameters that appeared to be the best at our search area (using gird-search method).

*A. Collision Counts*

In this section we report the time takes for each method to have a collision in its first seven encounters to the unseen (test) data. In the ideal case (ideal training), these values should be zeros, because the trained agent is expected to avoid collisions, however, due to the continuity of the state space it is highly possible that the agent faces a scenario which it has no idea about its safety.

*Table 4. Time to Collision*

|  | Traditional RL | Constrained Q | Q-masking |
|---|---|---|---|
| Test 1 | 25 | 33 | - |
| Test 2 | 31 | 38 | 22 |
| Test 3 | 18 | - | - |
| Test 4 | - | 37 | - |
| Test 5 | 16 | - | 36 |
| Test 6 | - | 34 | - |
| Test 7 | 24 | - | - |

It is noticeable in the *Table4* that the Q-masking has been the safest method among the others (had only two collisions), actually the low number of collisions resulted in Q-masking account for the heading angle which has not been considered in this algorithm. The notable thing is the better performance of constrained-Q versus traditional-RL even when its constraints are the linearized versions of the original non-linear ones.

The reason that we did not provide the results for our proposed robust Q-masking method in this table, is that it never encountered a collision in that seven episodes.

*B. Convergence time*

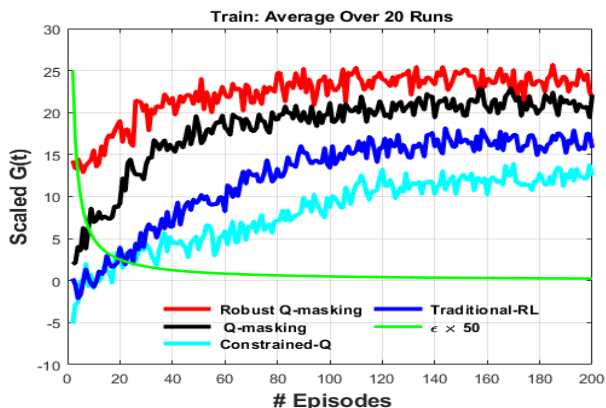

Figure 7: The scaled average reward and Epsilon for the train phase demonstrating the convergence time for each method.

In this section the introduced methods in Section IV are compared in terms of the speed to achieve their corresponding steady-state result, regardless of the quality of this final value. So, the vertical values are scaled in order to better showcase the comparison.

As it is noticeable from the *Fig7*, the proposed robust Q-masking has been the quickest method to learn its optimal behavior, immediately followed by Q-masking. Traditional-RL and constrained-Q-learning stand in the third and fourth place respectively.

*C. Quality of steady-states*

In this section, four methods of Section IV are compared in terms of average total reward. The reason that we did not begin this section with this comparison, is that the quality of reward is not what the only thing that matters in RL-based autonomous navigation. In fact, in autonomous navigation, "speedy driving" worth nothing if you encounter collisions from time to time.

Moreover, it should be noted that the below figure demonstrates the performance of each method averaged in times before having collision over 20 episodes, we believe this makes the comparison more reasonable because otherwise the proposed robust Q-masking will always outperform the traditional-RL, because it never encounters collisions, so it lasts more and gains more.

As it is noticeable from the above the Fig6, traditional- RL has performed better than other methods in the long run. However, the gap is not huge, and it is immediately followed by the Q-masking and proposed robust Q-masking. As for now, it is not clear to authors why the constrained Q-learning has performed so weak, we reviewed it repeatedly to make sure it is implemented correctly.

We believe the reason that traditional RL- performed better than other methods is that it is actually less conservative than them. It never takes into account the possible sudden lane changing of the nearby vehicles or their possible emergence braking. We believe this gap may further increase in the case of more lanes in the environment or more adjacent vehicles.

## VII. CONCLUSION

In this paper we investigated the problem of safe decision-making for autonomous navigation in the RL framework. We formulated the problem and then show that in contrast to common RL tasks, it is possible to make sure about the safety of the actions a priori. We reviewed different safe RL-based methods available in the literature. Taking the Q-masking as the focal point, we pointed out its disadvantageous and proposed robust Q-making as a modification with novel Q-function shape and inclusion of the heading angle. The implementation results confirm the improved sample efficiency of the proposed method.

## VIII. FUTURE WORK

This work can be followed in different ways. One can extend the action space to continuous space (e.g., the exact value for the acceleration) instead of discrete Meta actions in order to represent more realistic view of the problem.

Another interesting approach can classify the autonomous navigation task in a unified framework (i.e., inspection, and decision-making at the same time, see [27]). We also aim to measure the expense of our robust decision-making approach and compare that with the proposed state-belief methods in the literature, we are aware that out approach might be too conservative (sub-optimal) mainly due to inclusion of possible abrupt change in the heading angle of nearby vehicles (which will not happen most of the times), but we need to know about its sub-optimality in different scenarios. Another interesting thing is to repeat the implementations shown in Fig.6, but this time with the presence of one deterministic optimal policy to compare the methods in Section IV in terms of global optimality, because there are lots of suboptimal solution for the problem of autonomous driving.

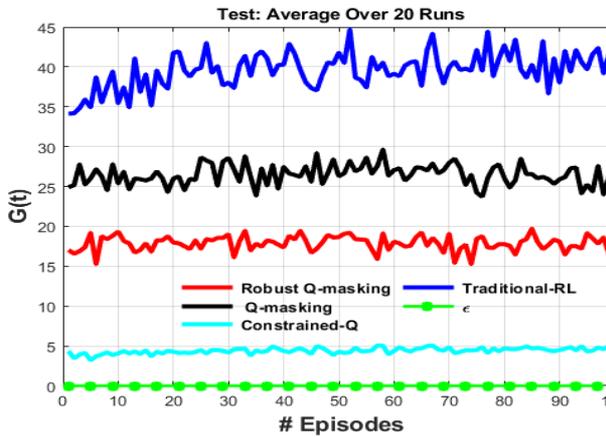

Figure 6: The average total reward for each method before encountering a collision

REFERENCES

[1] Nilsson, Nils J. Shakey the robot. SRI INTERNATIONAL MENLO PARK CA, 1984.
[2] Urmson, Chris, Joshua Anhalt, Drew Bagnell, Christopher Baker, Robert Bittner, M. N. Clark, John Dolan et al. "Autonomous driving in urban environments: Boss and the urban challenge." Journal of Field Robotics 25, no. 8 (2008): 425-466.
[3] Ziegler, Julius, Philipp Bender, Thao Dang, and Christoph Stiller. "Trajectory planning for Bertha—A local, continuous method." In *2014 IEEE intelligent vehicles symposium proceedings*, pp. 450-457. IEEE, 2014.
[4] Cosgun, Akansel, Lichao Ma, Jimmy Chiu, Jiawei Huang, Mahmut Demir, Alexandre Miranda Anon, Thang Lian, Hasan Tafish, and Samir Al-Stouhi. "Towards full automated drive in urban environments: A demonstration in gomentum station, california." In *2017 IEEE Intelligent Vehicles Symposium (IV)*, pp. 1811-1818. IEEE, 2017.
[5] Shalev-Shwartz, Shai, Shaked Shammah, and Amnon Shashua. "Safe, multi-agent, reinforcement learning for autonomous driving." arXiv preprint arXiv:1610.03295 (2016).
[6] Kümmerle, Rainer, Michael Ruhnke, Bastian Steder, Cyrill Stachniss, and Wolfram Burgard. "A navigation system for robots operating in crowded urban environments." In 2013 IEEE International Conference on Robotics and Automation, pp. 3225-3232. IEEE, 2013.
[7] Algorithms for Robust Autonomous Navigation in Human Environments, PhD thesis by by Michael F. Everett, MIT 2020.
[8] Montremerlo, M., J. Beeker, S. Bhat, and H. Dahlkamp. "The stanford entry in the urban challenge." *Journal of Field Robotics* 7, no. 9 (2008): 468-492.
[9] Hoel, Carl-Johan, Katherine Driggs-Campbell, Krister Wolff, Leo Laine, and Mykel J. Kochenderfer. "Combining planning and deep reinforcement learning in tactical decision making for autonomous driving." *IEEE Transactions on Intelligent Vehicles* 5, no. 2 (2019): 294-305.
[10] Mirchevska, Branka, Christian Pek, Moritz Werling, Matthias Althoff, and Joschka Boedecker. "High-level decision making for safe and reasonable autonomous lane changing using reinforcement learning." In *2018 21st International Conference on Intelligent Transportation Systems (ITSC)*, pp. 2156-2162. IEEE, 2018.
[11] Ulbrich, Simon, and Markus Maurer. "Towards tactical lane change behavior planning for automated vehicles." In 2015 IEEE 18th International Conference on Intelligent Transportation Systems, pp. 989- 995. IEEE, 2015.
[12] Sutton, Richard S., and Andrew G. Barto. Reinforcement learning: An introduction. MIT press, 2018.
[13] Mnih, Volodymyr, Koray Kavukcuoglu, David Silver, Andrei A. Rusu, Joel Veness, Marc G. Bellemare, Alex Graves et al. "Human-level control through deep reinforcement learning." nature 518, no. 7540 (2015): 529-533.
[14] Tehrani, Hossein, Quoc Huy Do, Masumi Egawa, Kenji Muto, Keisuke Yoneda, and Seiichi Mita. "General behavior and motion model for automated lane change." In 2015 IEEE Intelligent Vehicles Symposium (IV), pp. 1154-1159. IEEE, 2015.
[15] Sharifzadeh, Sahand, Ioannis Chiotellis, Rudolph Triebel, and Daniel Cremers. "Learning to drive using inverse reinforcement learning and deep q-networks." arXiv preprint arXiv:1612.03653 (2016).
[16] Bojarski, Mariusz, Davide Del Testa, Daniel Dworakowski, Bernhard Firner, Beat Flepp, Prasoon Goyal, Lawrence D. Jackel et al. "End to end learning for self-driving cars." arXiv preprint arXiv:1604.07316 (2016).
[17] Xu, Huazhe, Yang Gao, Fisher Yu, and Trevor Darrell. "End-to-end learning of driving models from large-scale video datasets." In Proceedings of the IEEE conference on computer vision and pattern recognition, pp. 2174-2182. 2017.
[18] Chen, Chenyi, Ari Seff, Alain Kornhauser, and Jianxiong Xiao. "Deepdriving: Learning affordance for direct perception in autonomous driving." In Proceedings of the IEEE International Conference on Computer Vision, pp. 2722-2730. 2015.
[19] Shalev-Shwartz, Shai, Shaked Shammah, and Amnon Shashua. "Safe, multi-agent, reinforcement learning for autonomous driving." arXiv preprint arXiv:1610.03295 (2016).
[20] Bai, Haoyu, David Hsu, and Wee Sun Lee. "Integrated perception and planning in the continuous space: A POMDP approach." The International Journal of Robotics Research 33, no. 9 (2014): 1288-1302.
[21] LeCun, Yann, Yoshua Bengio, and Geoffrey Hinton. "Deep learning." nature 521, no. 7553 (2015): 436-444.
[22] Mnih, Volodymyr, Koray Kavukcuoglu, David Silver, Andrei A. Rusu, Joel Veness, Marc G. Bellemare, Alex Graves et al. "Human-level control through deep reinforcement learning." nature 518, no. 7540 (2015): 529-533.
[23] https://highway-env.readthedocs.io/en/latest/index.html
[24] Mukadam, Mustafa, Akansel Cosgun, Alireza Nakhaei, and Kikuo Fujimura. "Tactical decision making for lane changing with deep reinforcement learning." (2017).
[25] Mirchevska, Branka, Christian Pek, Moritz Werling, Matthias Althoff, and Joschka Boedecker. "High-level decision making for safe and reasonable autonomous lane changing using reinforcement learning." In 2018 21st International Conference on Intelligent Transportation Systems (ITSC), pp. 2156-2162. IEEE, 2018.
[26] Li, Xin, Xin Xu, and Lei Zuo. "Reinforcement learning based overtaking decision-making for highway autonomous driving." In 2015 Sixth International Conference on Intelligent Control and Information Processing (ICICIP), pp. 336-342. IEEE, 2015.
[27] Sallab, Ahmad EL, Mohammed Abdou, Etienne Perot, and Senthil Yogamani. "Deep reinforcement learning framework for autonomous driving." Electronic Imaging 2017, no. 19 (2017): 70-76.
[28] Shalev-Shwartz, Shai, Shaked Shammah, and Amnon Shashua. "Safe, multi-agent, reinforcement learning for autonomous driving." arXiv preprint arXiv:1610.03295 (2016).
[29] Sallab, Ahmad EL, Mohammed Abdou, Etienne Perot, and Senthil Yogamani. "Deep reinforcement learning framework for autonomous driving." Electronic Imaging 2017, no. 19 (2017): 70-76.
[30] Ulbrich, Simon, and Markus Maurer. "Towards tactical lane change behavior planning for automated vehicles." In 2015 IEEE 18th International Conference on Intelligent Transportation Systems, pp. 989-995. IEEE, 2015.
[31] Ulbrich, Simon, and Markus Maurer. "Probabilistic online POMDP decision making for lane changes in fully automated driving." In 16th International IEEE Conference on Intelligent Transportation Systems (ITSC 2013), pp. 2063-2067. IEEE, 2013.
[32] Hoel, Carl-Johan, Krister Wolff, and Leo Laine. "Automated speed and lane change decision making using deep reinforcement learning." In 2018 21st International Conference on Intelligent Transportation Systems (ITSC), pp. 2148-2155. IEEE, 2018.
[33] Pek, Christian, Peter Zahn, and Matthias Althoff. "Verifying the safety of lane change maneuvers of self-driving vehicles based on formalized traffic rules." In *2017 IEEE Intelligent Vehicles Symposium (IV)*, pp. 1477-1483. IEEE, 2017.
[34] Economic Commission for Europe: Inland Transport Committee, "Vienna Convention on Road Traffic," 1968. [Online]. Available: http://www.unece.org/fileadmin/DAM/trans/conventn/crt1968e.pdf.
[35] Jin, Chi, Zeyuan Allen-Zhu, Sebastien Bubeck, and Michael I. Jordan. "Is Q-learning provably efficient?." In Advances in neural information processing systems, pp. 4863-4873. 2018.
[36] Deisenroth, M., and Rasmussen CE PILCO. "A model-based and data-efficient approach to policy search." In *Proceedings of the 28th International Conference on Machine Learning*, pp. 465-472.
[37] Schulman, John, Sergey Levine, Pieter Abbeel, Michael Jordan, and Philipp Moritz. "Trust region policy optimization." In *International conference on machine learning*, pp. 1889-1897. 2015.
[38] Fortunato, Meire, Mohammad Gheshlaghi Azar, Bilal Piot, Jacob Menick, Ian Osband, Alex Graves, Vlad Mnih et al. "Noisy networks for exploration." *arXiv preprint arXiv:1706.10295* (2017).
[39] Safe Q-Learning Method Based on Constrained Markov Decision Processes.
[40] Ge, Yangyang, Fei Zhu, Xinghong Ling, and Quan Liu. "Safe Q-Learning Method Based on Constrained Markov Decision Processes." *IEEE Access* 7 (2019): 165007-165017.
[41] Bušić, Ana, and Sean Meyn. "Action-constrained Markov decision processes with Kullback-Leibler cost." *arXiv preprint arXiv:1807.10244* (2018).
[42] Borkar, Vivek S. "An actor-critic algorithm for constrained Markov decision processes." *Systems & control letters* 54, no. 3 (2005): 207-213.
[43] Achiam, Joshua, David Held, Aviv Tamar, and Pieter Abbeel. "Constrained policy optimization." *arXiv preprint arXiv:1705.10528* (2017).